\documentclass[letterpaper, 10 pt, conference]{ieeeconf}  

\IEEEoverridecommandlockouts                              

\usepackage{graphics} 
\usepackage{epsfig} 
\usepackage{mathptmx} 
\usepackage{times} 
\usepackage{amsmath} 
\usepackage{amssymb}  
\usepackage{graphicx}

\usepackage[breaklinks=true,bookmarks=false]{hyperref}

\title{\LARGE \bf
Double Refinement Network for Efficient Monocular Depth Estimation
}

\author{Nikita Durasov$^{1}$, Mikhail Romanov$^{1}$, Valeriya Bubnova$^{1}$, Pavel Bogomolov$^{1}$, Anton Konushin$^{1}$
\thanks{$^{1}$Samsung AI Center Moscow, \{n.durasov, m.romanov, v.bubnova, p.bogomolov, a.konushin\}@samsung.com    
}%
}

\begin{document}

\maketitle
\thispagestyle{empty}
\pagestyle{empty}

\begin{abstract}
Monocular depth estimation is the task of obtaining a measure of distance for each pixel using a single image. It is an important problem in computer vision and is usually solved using neural networks. Though recent works in this area have shown significant improvement in accuracy, the state-of-the-art methods tend to require massive amounts of memory and time to process an image. The main purpose of this work is to improve the performance of the latest solutions with no decrease in accuracy. To this end, we introduce the Double Refinement Network architecture. The proposed method achieves state-of-the-art results on the standard benchmark RGB-D dataset NYU Depth v2, while its frames per second rate is significantly higher (up to 18 times speedup per image at batch size 1) and the RAM usage per image is lower.
\end{abstract}

\section{INTRODUCTION}

In computer vision, depth estimation is the task of obtaining a measure of distance for each pixel of an image. It has a wide array of applications in consumer electronics (for example, in mobile photography) as well as in other computer vision tasks (simultaneous localization and mapping, visual odometry, to name a few). One of the most challenging settings in this field is monocular depth estimation, i.e. retrieving depth from a single image. Unlike other scenarios, such as stereo depth estimation, it does not require any additional equipment, and therefore can be applied in any devices with a camera, e.g. mobile phones, augmented reality headsets, indoor robots. In our work, we focus on monocular depth estimation exclusively.

We consider the regression problem of point-wise depth reconstruction. More specifically, we want to train a model $f(x)$: $\mathbb{R}_{H \times W \times 3}\to \mathbb{R}_{H \times W \times 1}$ to predict the distance from the camera to the corresponding object for each pixel.

\begin{table*}[htb]
\caption{Network predictions compared to the ground truth. From top to bottom: input, network depth map prediction, ground truth depth map}
\label{fig:visual_results}
\begin{center}
\begin{tabular}{cccc}
             \includegraphics[width=0.2\textwidth]{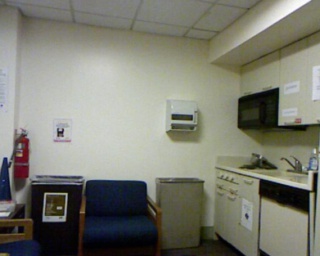} & \includegraphics[width=0.2\textwidth]{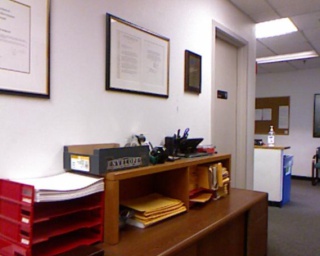} & \includegraphics[width=0.2\textwidth]{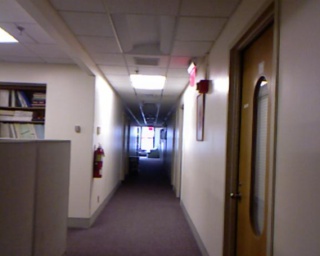} & \includegraphics[width=0.2\textwidth]{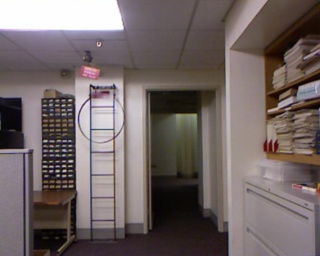}\\
             \includegraphics[width=0.2\textwidth]{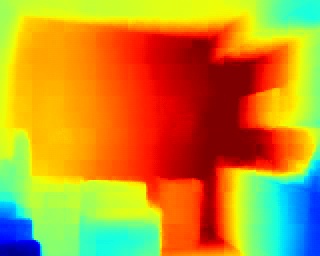} & \includegraphics[width=0.2\textwidth]{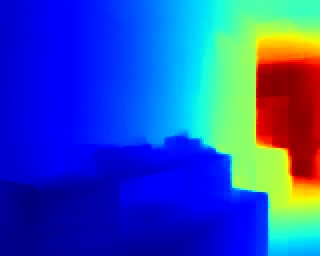} & \includegraphics[width=0.2\textwidth]{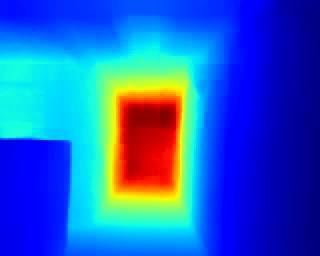} & \includegraphics[width=0.2\textwidth]{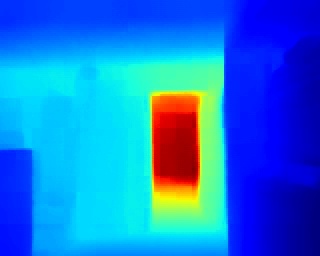}\\
             \includegraphics[width=0.2\textwidth]{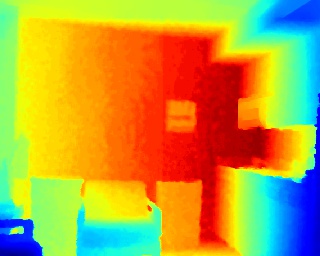} & \includegraphics[width=0.2\textwidth]{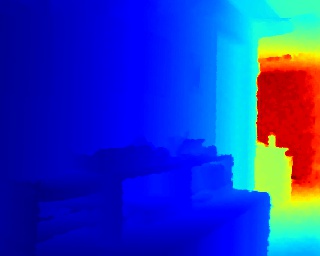} & \includegraphics[width=0.2\textwidth]{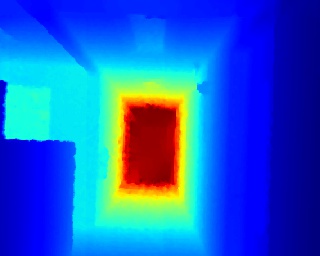} & \includegraphics[width=0.2\textwidth]{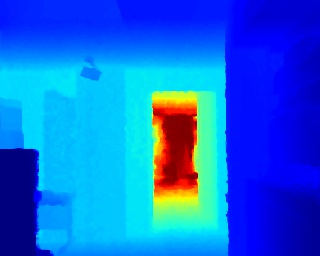}\\
\end{tabular}
\end{center}
\end{table*}

Recent approaches employ deep neural networks to solve this task and show significant improvements in quality over classical methods. However, their time and memory requirements for processing a single image are very high, which renders them unusable for production purposes, especially in portable devices. One of the most computationally expensive parts of these networks seems to be the bilinear interpolation of feature maps to the input size since the resulting tensor typically has very high dimensions. We propose a network architecture, called Double Refinement Network, which performs the interpolation iteratively while reducing the number of channels on each level. We evaluate our model on the standard benchmark RGB-D dataset NYU Depth v2 and show noticeable performance gain without compromising accuracy.

Firstly, we overview the recent work on the topic of monocular depth estimation. After that, we describe our architecture and experiments. Finally, we discuss the results and summarize our contribution.

\section{RELATED WORK OVERVIEW}

\subsection{Classical methods}
Depth estimation has been actively researched for over a decade. First methods did not rely on deep learning and instead utilized more algorithmic approaches. 

Saxena et al. \cite{saxena2009make3d} used Markov Random Field to divide an image into superpixels, and reconstructed the 3D structure of the image by estimating the position and orientation of the 3D surfaces that the superpixels represented. Several geometry-based algorithms for indoor depth estimation including \cite{hedau2010thinking} and \cite{gupta2010estimating} were based on a number of assumptions about room shapes and the usual object placement in such rooms. \cite{gupta2010estimating} used SVM to compute a feature vector for each surface. \cite{hoiem2005geometric} presented a system that modeled geometric classes depending on the orientation of a physical scene and built the geometric structure progressively. Geometry-based approaches worked with certain constraints and were relevant only for specific types of images.

\begin{figure*}[htb]
    \begin{center}
    \centering
    \includegraphics[width=1.0\textwidth]{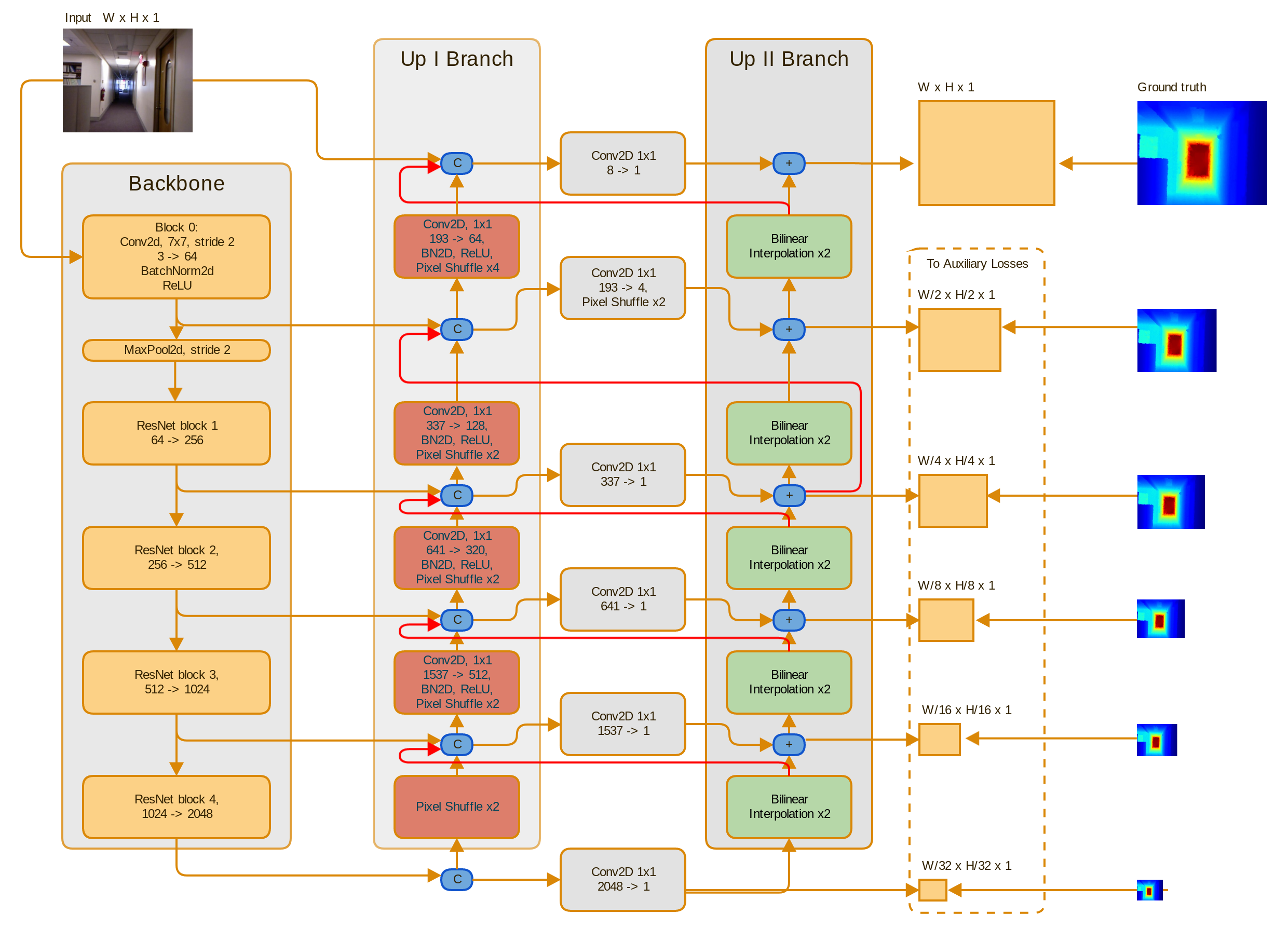}
    \caption{Double Refinement Network architecture. The downsampling (yellow) branch, represents the backbone. The figure shows ResNet $\geq$ 50 as a backbone, but the logic for the other pre-trained networks remains the same. 
    The network contains two upsampling branches (red and green). Red branch (\texttt{up I}) merges the high-level features with the low-level features and upsamples them with pixel shuffle. 
    The green branch (\texttt{upII}) outputs intermediate depth maps on each scale, which enables us to compute the loss on every level and supports the training of the lower layers. To propagate the predictions to higher levels, we use bilinear interpolation. The red arrows are referred to as the diagonal connections in the text.}
    \label{fig:w_net}
    \end{center}
\end{figure*}

\subsection{Deep Learning Methods}

Most recent works on depth reconstruction utilize Deep Convolutional Neural Networks (DCNN) \cite{eigen2015predicting, krizhevsky2012imagenet, kuznietsov2017semi, laina2016deeper, liu2016learning}. It is common practice to use pre-trained layers (e.g. VGG \cite{simonyan2014very}, SENet \cite{hu2018squeeze} or ResNet \cite{he2016deep}) for feature extraction. Another noticeable trend is exploiting encoder-decoder architectures.
In such architectures, convolutional and max-pooling layers usually form the encoder and upsampling layers often play the role of the decoder.
Here we overview some of the latest works with high benchmark results. 

In \cite{gurram2018monocular} Gurram et al. propose a solution which includes training a model on two heterogeneous datasets, one with depth and the other with semantic map labels. Common NN layers are trained alternately on data from either of the datasets. In particular, the network consists of two flows that can be used for getting the depth or the semantic map of an image respectively.

Fu et al. \cite{fu2018deep} formulate monocular depth estimation as an ordinal regression problem, i.e. they discretize depth. Their architecture includes a dense feature extractor (convolutions and dense layers), a scene understanding modular (a concatenation of dilated convolutions, a full-image encoder, and convolutional layers) and an ordinal regression block.

He et al. \cite{he2018learning} make a point that without prior knowledge about the focal length of the camera, depth prediction can be ambiguous. They suggest that using datasets with various focal lengths may be helpful in handling this issue. The proposed method involves generating multiple images with different focal lengths from one fixed-focal-length image. Thus, the model is trained on a varying-focal-length dataset. Focal lengths are given to the network in addition to images.

Spek et al. \cite{spek2018cream} train an encoder-decoder-solver architecture and approximate the encoder and decoder with their less computationally expensive versions. As a result, their architecture allows real-time depth estimation.

RefineNet \cite{lin2017refinenet} is a segmentation network which can also be used for depth estimation. It is similar to U-Net but is based on ResNet. One of its advantages is a high FPS rate compared to other state-of-the-art architectures.

Hu et al. \cite{hu2018revisiting} interpolate the ResNet feature maps of all the encoder layers to the size of the input and concatenate them. The resulting feature map is then passed into a custom decoder. This method, which we refer to as RSIDE, achieves state-of-the-art results and is the baseline for our work.

\section{METHOD}

We improve upon the baseline by making the following changes:
\begin{itemize}
    \item Substitute bilinear interpolation of the feature maps to the size of the input with iterative pixel shuffle upsampling (in \texttt{upI})
    
    \item Add another branch for intermediate depth maps (\texttt{upII}) with auxiliary losses applied to them
\end{itemize}

We call the resulting architecture Double Refinement Network. As we show further, this architecture is significantly more computationally efficient than the baseline.

\subsection{Relation to the State-of-the-Art Architectures}

The current state-of-the-art architectures have high requirements for computational resources and memory.
Some of the architectures (such as in \cite{hu2018revisiting}) use bilinear interpolation
of all the deep activation maps to the input size. This procedure is very time- and memory-consuming, which leads to a low FPS rate.

PSPNet \cite{zhao2017pyramid} and DeepLab v3 \cite{chen2017rethinking}, \cite{chen2018encoder}, which can be applied to the depth estimation problem, use computationally expensive dilated convolutions.

In the proposed architecture, we replace dilated convolutions and bilinear interpolation of high-dimensional activation maps with iterative pixel shuffle upsampling. However, such changes may cause a decrease in accuracy.

To mitigate this, we propose a double-refinement framework. We start by obtaining a low-scale depth estimation, which we then iteratively improve, using the high-level backbone features. On each scale, we compute the correction term from the previous approximation and the lower-level backbone features. This requires less information and enables us to only upsample feature maps between consecutive levels, saving time and memory needed to interpolate them to the input size. We add this correction term to the upsampled depth estimation from the previous level, producing a new estimation at twice the resolution. The final output is of the same size as the input.

\subsection{Architecture Description}

Fig. \ref{fig:w_net} shows the architecture of the proposed network.

The network consists of one downsampling branch (backbone) and two upsampling branches. 

In this paper, we use the following notation: $\texttt{down}_i$ is the output of the \texttt{down} branch on the level $i$. For the upsampling branches we count from the top, $i$-th blocks of the \texttt{upI} and \texttt{upII} branches we denote as \texttt{upI}$_i$ and \texttt{upII}$_i$ correspondingly. Besides, BI($\cdot$) stands for bilinear interpolation by a factor of $2$.

We use a pre-trained backbone network, experimenting with different architectures: ResNet-50, ResNet-152 \cite{he2016deep}, DenseNet-161 \cite{huang2017densely} and SENet-154 \cite{hu2018squeeze}. We remove the last fully connected layer and its respective average pooling layer from the pre-trained networks. 

The backbone network is split into several blocks
(\texttt{layer 0,1,2,3} and \texttt{4}).
Note that the downsample factor of \texttt{layer 0} is 4, while for the other blocks, except \texttt{layer 1}, it is 2. We denote the output of \texttt{layer i} as $\texttt{down}_{i+1}$.

The upsampling process is similar to U-Net \cite{ronneberger2015u} and RefineNet \cite{lin2017refinenet}. To start describing it, we substitute $\texttt{down}_5$ with $\texttt{upI}_5$ for convenience. $\texttt{upI}_5$ contains the high-level features extracted from the input.

To get the first approximation of the depths $\texttt{upII}_5$
we apply \texttt{Conv2D} 1 $\times$ 1 to $\texttt{upI}_5$.

To compute $\texttt{upI}_i$, we perform the following operations  with $\texttt{upI}_{i+1}$:

\begin{itemize}
  \item \texttt{Conv2d 1$\times$1} (except $\texttt{upI}_{5}$ block)
  \item \texttt{BatchNorm2D} (except $\texttt{upI}_{5}$)
  \item \texttt{ReLU} (except $\texttt{upI}_{5}$)
  \item \texttt{PixelShuffle $\times$2} (except $\texttt{upI}_{1}$; for $\texttt{upI}_{0}$ -- \texttt{PixelShuffle $\times 4$}).
\end{itemize}

\begin{table*}[h]
\caption{Comparison of our network with different backbones against other networks. 
We should note that with the ResNet-50 backbone our network performs worse than the baseline. 
We are able to narrow this gap with architectural changes (see experiments in \ref{more_exp}).}
\label{fig:table_of_results}
    \begin{center}
    \begin{tabular}{cccccc}
    \hline
    & RMSE $\downarrow$ & Log10 $\downarrow$ & $\delta_{1.25}$ $\downarrow$ & $\delta_{1.25 ^ 2}$ $\uparrow$ & $\delta_{1.25^3}$ $\uparrow$ \\
    \hline
\hline

\hline

Lauina et al. \cite{laina2016deeper} & 0.573 & 0.055 & 0.811 & 0.953 & 0.988 \\
Fu et al. \cite{fu2018deep} & \textbf{0.509} & 0.051 & 0.828 & 0.965 & 0.992 \\
Spek et al. \cite{spek2018cream} & 0.687 & 0.161 & 0.704 & 0.917 & 0.977 \\
Nekrasov et al. \cite{nekrasov2018real} & 0.565 & - & 0.790 & 0.955 & 0.990 \\
RSIDE \cite{hu2018revisiting} ResNet-50 (paper) & 0.555 & 0.054 & 0.843 & 0.968 & 0.991 \\
RSIDE DenseNet-161 (paper) & 0.544 & 0.053 & 0.855 & 0.972 & 0.993 \\
RSIDE SENet-154 (paper) & 0.530 & \textbf{0.050} & 0.866 & \textbf{0.975} & 0.993 \\
RSIDE SENet-154 (reproduced) & 0.574 & 0.052 & 0.845 & 0.967 & 0.991 \\

\hline
(ours) DRNet ResNet-50  & 0.587 & 0.061 & 0.810 & 0.961 & 0.990 \\
(ours) DRNet ResNet-152 & 0.528 & 0.050 & 0.866 & 0.973 & 0.993 \\
(ours) DRNet DenseNet-161 & 0.534 & 0.052 & 0.865 & 0.974 & \textbf{0.994} \\
(ours) DRNet SeNet-154 & 0.527 & \textbf{0.050} & \textbf{0.868} & \textbf{0.975} & 0.993 \\
\hline
\end{tabular}
\end{center}
\end{table*}

The output is concatenated with $BI(\texttt{upII}_{i + 1})$, which is a lightweight operation due to the small size of intermediate outputs.

The second upsampling branch of the network \texttt{upII} is inspired by Fourier Transform. We make a coarse prediction and then, ascending from the bottom to the top, we make corrections to this prediction. Each level increases the side of the output by a factor of $2$.

In particular, the depth approximation $\texttt{upII}_{i}$ is the sum of $BI(\texttt{upII}_{i + 1})$ and the correction term. To compute the latter, we concatenate $\texttt{down}_i$, $BI(\texttt{upII}_{i + 1})$ and $\texttt{upI}_{i}$ and apply \texttt{Conv2D 1$\times$1} to the result.

\begin{table*}[h]
\caption{Comparison of our network with different backbones against other models in FPS and memory consumption.
The FPS rates are measured using Tesla P40 GPU. In this table, "BS" stands for the batch size. }
\label{fig:speed_comparison}
    \begin{center}
    \begin{tabular}{ccccccccccc}
    \hline
    & FPS (BS 1) $\uparrow$ & FPS (max BS) $\uparrow$ & net RAM $\downarrow$ & RAM / img $\downarrow$ \\
    \hline
\hline

\hline
Nekrasov et al. \cite{nekrasov2018real} & 78 & - & - & - \\
RSIDE ResNet-50 (reproduced) & 2 & 2 (BS 1) & 793 Mb & 21 Gb \\
RSIDE DenseNet-161 (reproduced) & 1.7 & 1.7 (BS 1) & 853 Mb & 22 Gb  \\
RSIDE SENet-154 (reproduced) & 1.5 & 1.5 (BS 1) & 1.7 Gb & 21.9 Gb \\

\hline
(ours) DRNet ResNet-50 & 36 & 69 (BS 36) & 757 Mb & 2 Gb\\
(ours) DRNet ResNet-152 & 25 & 55 (BS 19) & 1067 Mb & 2.9 Gb \\
(ours) DRNet DenseNet-161 & 17 & 33 (BS 14) & 839 Mb & 3.2 Gb \\
(ours) DRNet SeNet-154 & 6 & 13 (BS 9) & 1421 Mb & 4 Gb \\
\hline
\end{tabular}
\end{center}
\end{table*}

\subsection{Loss Function}

Depth estimation requires not only pixel-wise accuracy but also spatially 
coherent result. That is the reason why depth estimation models tend to incorporate depth gradient and normals into their loss functions. The loss we use consists of three parts:

\begin{equation}
    L_{i} = l_{i}^{depth} + l_{i}^{grad} + l_{i}^{normal},
\end{equation}

\begin{equation}
    l_{i}^{depth} = \cfrac{1}{n} \sum_{j=0}^n F(e_j),
\end{equation}

\begin{equation}
    l_{i}^{grad} = \cfrac{1}{n} \sum_{j=0}^n \left( F(\nabla_x(e_j)) + F(\nabla_y(e_j)) \right),
\end{equation}

where $F(x) = \ln(x + \alpha)$, $\alpha > 0$ and $e_j = \|d_j - g_j\|_1$ ($d$ and $g$ standing for predicted and ground truth depth maps respectively),

\begin{equation}
    l_{i}^{normal} = \cfrac{1}{n} \sum_{j=0}^n \left( 1 - \cfrac{\langle n_j^d, n_j^g \rangle}{\sqrt{\langle n_j^d, n_j^d \rangle} \sqrt{\langle n_j^g, n_j^g \rangle}} \right),
\end{equation}

with $n^d$ and $n^g$ standing for predicted and ground truth normals correspondingly:

\begin{equation}
    n_i^d = [-\nabla_x(d_i), -\nabla_y(d_i), 1]^T,
\end{equation}

\begin{equation}
    n_i^g = [-\nabla_x(g_i), -\nabla_y(g_i), 1]^T.
\end{equation}

These loss functions are described in-depth in \cite{hu2018revisiting}, we apply them without changes.

\subsection{Auxiliary Loss Functions}

We treat the intermediate output on $i$-th level as a downsampled depth map and calculate the loss between it and the target, downsampled to $\frac{1}{2^i}$ of its original size.

Thus, the overall loss function is computed as:
\begin{equation}
    L = \sum_{i=0}^5 L_{i}.
\end{equation}

We have also tried upsampling the level-specific result using bilinear interpolation and found that it works worse than downsampling the target.

\section{EXPERIMENTS}

We conduct our experiments on the NYU Depth v2 dataset.
We use the following common metrics to compare our model to the state-of-the-art models:

\begin{itemize}
\item Root mean squared error (RMSE): 
\begin{equation}
    \sqrt{\frac{1}{|N|} \sum_{i \in N} |d_i - g_i|^2}
\end{equation}
\item Threshold ($\delta$):
\begin{equation}
    \text{\% of $d_i$ s.t.} \max{\left(\frac{d_i}{g_i}, \frac{g_i}{d_i}\right) < t},
\end{equation}
where
$t \in \{1.25, 1.25^2, 1.25^3\}$
\end{itemize}

\begin{table}[h]
\caption{Experiments with additional improvements to the architecture}
\label{fig:resnet50_experiments}
    \begin{center}
    \begin{tabular}{lccccc}
    \hline
    & RMSE & $\delta_{1.25}$ & $\delta_{1.25 ^ 2}$ & $\delta_{1.25^3}$ &FPS\\
    \hline
\hline

\hline
DRNet ResNet-50 & 0.587 & 0.810 & 0.961 & 0.990 & 36 \\
+ freeze backbone on train & 0.588 & 0.812 & 0.961 & 0.990 & \\
+ Guided Filter & 0.564 & 0.836 & 0.967 & 0.992 & 30 \\
+ kernel size 3 in upI-upII & 0.582 & 0.818 & 0.961 & 0.991 &     \\
+ kernel size 5 in upI-upII & 0.579 & 0.825 & 0.964 & 0.990 & 37 \\
\hline
DRNet ResNet-152 & 0.528 & 0.866 & 0.961 & 0.990 & \\
+ Guided Filter & 0.537 & 0.867 & 0.973 & 0.993 & \\
+ kernel size 5 in upI-upII & 0.532 & 0.864 & 0.973 & 0.993 & \\
\hline
\end{tabular}
\end{center}
\end{table}

Tables \ref{fig:table_of_results}, \ref{fig:speed_comparison} show that our proposed architecture achieves results comparable to the baseline while providing up to 35 times faster inference and requiring up to 10 times less memory per image. In real-time scenarios (batch size 1), the inference FPS rate is up to 18 times higher than the baseline's. The visual results can be found in Table \ref{fig:visual_results}.


\subsection{Implementation details}


We train the model using the pre-trained backbone networks from the torchvision package \cite{paszke2017automatic}. We use Adam (amsgrad modification) with learning rate $10^{-4}$, weight decay $10^{-4}$ and the default betas.

\subsection{More experiments}
\label{more_exp}


Table \ref{fig:table_of_results} demonstrates that the accuracy of the ResNet-50-based network is lower than of the architectures with backbones of higher capacity. We experiment with different changes to the ResNet-50-based model with the intention of transferring the improvements to the more capacious architectures. Table \ref{fig:resnet50_experiments} represents the results of these experiments.

\subsubsection{Frozen weights} We train the model with frozen backbone weights. No improvements are observed in several experimental settings.

\subsubsection{Guided Filter} We substitute bilinear interpolation of intermediate outputs in $\texttt{upII}$ with the Guided Filter upsampling method proposed in \cite{wu2017fast}. This change provides an increase in accuracy while retaining similar performance.

\subsubsection{Correction term layers} We study the importance of the layers that produce the correction term for the intermediate outputs (grey blocks in Fig. \ref{fig:w_net}). We increase the receptive field of the layers by changing the kernel size. This results in a higher score and does not slow down the inference.

While introducing these changes to the architecture with the ResNet-50 backbone improves the metrics significantly, they do not provide substantial benefits to the models with more capacious backbones.


\subsection{Ablation Studies}

We introduce several changes to the baseline.
To ensure that each of them contributes to the result, we conduct element reasoning experiments. The results are shown in Table \ref{fig:elements_reasoning}. From them, we infer that the second upsampling branch gives the most noticeable improvement. The auxiliary losses slightly improve the metrics, while the diagonal connections appear to have no effect on the result and can be excluded from the network.

\begin{table}[h]
\caption{Results of experiments with exclusion of each of the proposed elements from Double Refinement Network with the ResNet-152 backbone.}
\label{fig:elements_reasoning}
    \begin{center}
    \begin{tabular}{cccccccccc}
    \hline
    & RMSE & $\delta_{1.25}$ & $\delta_{1.25 ^ 2}$ & $\delta_{1.25^3}$ \\
    \hline
\hline

\hline
 DRNet ResNet-152 & 0.528 & 0.866 & 0.972 & 0.993 \\
 no diagonal connections & 0.533 & 0.866 & 0.973 & 0.993 \\
 no auxiliary losses & 0.544 & 0.860 & 0.971 & 0.992 \\
 no second branch & 1.900 & 0.351 & 0.596 & 0.732 \\
\hline
\end{tabular}
\end{center}
\end{table}

\section{CONCLUSIONS}

In this paper, we introduced a new architecture for monocular depth estimation.
The network works significantly faster and uses 10 times less memory per image compared to the baseline while achieving state-of-the-art quality. It allows real-time inference at up to 36 FPS without network compression. 
We achieved this by replacing bilinear interpolation of feature maps to the input size with iterative refinement and applying the loss function to the intermediate outputs.

\addtolength{\textheight}{-12cm}   








{\small
\bibliographystyle{ieee}
\bibliography{egbib}
}

\end{document}